\documentclass[a4paper, 10pt, conference]{ieeeconf}      % Use this line for a4
\usepackage{FG2024}

\usepackage{graphicx}
\usepackage{amsmath}
\usepackage{amssymb}
\usepackage{booktabs}
\usepackage{amssymb}
\usepackage[normalem]{ulem}
\usepackage{multirow}
\usepackage{hhline}
\usepackage{hyperref}
\usepackage{textcomp}
\usepackage{wrapfig}
\usepackage{pifont}
\usepackage{float}
\usepackage{fancyhdr}

\usepackage{xurl}

\newcommand{\xmark}{\ding{55}}
\newcommand{\etal}{\textit{et al}. }
\usepackage[capitalize]{cleveref}
\crefname{section}{Sec.}{Secs.}
\Crefname{section}{Section}{Sections}
\Crefname{table}{Table}{Tables}
\crefname{table}{Tab.}{Tabs.}
\FGfinalcopy % *** Uncomment this line for the final submission

\def\FGPaperID{154} % *** Enter the FG2024 Paper ID here

\title{\LARGE \bf
EmoCLIP: A Vision-Language Method for Zero-Shot Video Facial Expression Recognition
}

\author{\parbox{16cm}{\centering
    {\large Niki Maria Foteinopoulou$^1$ and Ioannis Patras$^1$}\\
    {\normalsize
    $^1$ School of Electronic Engineering and Computer Science, \\ Queen Mary University of London, London, United Kingdom\\
    \thanks{This work is supported by EU H2020 AI4Media (No. 951911) and EPSRC DTP studentship (No. EP/R513106/1). This research utilised Queen Mary's Apocrita HPC facility, supported by QMUL Research-IT.\url{http://doi.org/10.5281/zenodo.438045}
    } % <-this % stops a space
}}}

\begin{document}

\ifFGfinal
\else
\author{Anonymous FG2024 submission\\ Paper ID \FGPaperID \\}
\pagestyle{plain}
\fi
\maketitle
\thispagestyle{fancy}

%%% END BLOCK HEADER AND COPYRIGHT NOTICE %%%

%%%%%%%%%%%%%%%%%%%%%%%%%%%%%%%%%%%%%%%%%%%%%%%%%%%%%%%%%%%%%%%%%%%%%%%%%%%%%%%%
\begin{abstract}

Facial Expression Recognition (FER) is a crucial task in affective computing, but its conventional focus on the seven basic emotions limits its applicability to the complex and expanding emotional spectrum. To address the issue of new and unseen emotions present in dynamic in-the-wild FER, we propose a novel vision-language model that utilises sample-level text descriptions (i.e. captions of the context, expressions or emotional cues) as natural language supervision, aiming to enhance the learning of rich latent representations, for zero-shot classification. To test this, we evaluate using zero-shot classification of the model trained on sample-level descriptions on four popular dynamic FER datasets. Our findings show that this approach yields significant improvements when compared to baseline methods. Specifically, for zero-shot video FER, we outperform CLIP by over 10\% in terms of Weighted Average Recall and 5\% in terms of Unweighted Average Recall on several datasets. Furthermore, we evaluate the representations obtained from the network trained using sample-level descriptions on the downstream task of mental health symptom estimation, achieving performance comparable or superior to state-of-the-art methods and strong agreement with human experts. Namely, we achieved a Pearson's correlation coefficient of up to 0.85 for schizophrenia symptom severity estimation, which is comparable to human experts' agreement. The code is publicly available on \url{https://github.com/NickyFot/EmoCLIP.git}.

\end{abstract}

%%%%%%%%%%%%%%%%%%%%%%%%%%%%%%%%%%%%%%%%%%%%%%%%%%%%%%%%%%%%%%%%%%%%%%%%%%%%%%%%
\section{Introduction}
\label{sec:intro}

Facial Expression Recognition (FER) is a primary task of affective computing, with several practical applications in Human-Computer Interaction~\cite{chowdary2021deep}, education~\cite{YADEGARIDEHKORDI2019103649} and mental health~\cite{foteinopoulou_learning_2022}, among others. To formalise the spectrum of human emotions, researchers have proposed several models. Ekman \& Friesen~\cite{ekman_facial_1978} propose six emotions (later seven with the addition of ``contempt'') as the basis of human emotional expression. This is the most widely accepted model for FER; however, human emotional experience is significantly more complex and varied than the seven basic categories, with up to 27 distinct categories for emotion reported in recent studies~\cite{cowen_self-report_2017}. A continuous arousal-valence scale has been proposed~\cite{russell_circumplex_1980} as an alternative to creating emotion categories; however, the scale is neither objective~\cite{foteinopoulou_estimating_2021} nor self-explanatory to human readers. Therefore, as more fine-grained definitions of emotion are proposed and the categorical models expand, there is also a need for automated systems to adjust to new definitions, unseen emotions and mental states or compound emotions. Zero-shot methodologies in Facial Emotion Recognition (FER) specifically tackle these challenges, providing effective solutions to unseen emotions and mental states.

Zero-shot Learning (ZSL) is a machine learning paradigm where a model can recognise and classify objects or concepts it has never been trained on by leveraging auxiliary information or attributes associated with those unseen classes~\cite{xian2018zero}. In the context of emotion recognition, ZSL has traditionally been achieved by some label semantic embedding to produce emotion prototypes, typically word2vec~\cite{zhan_zero-shot_2019, banerjee_learning_2022}. The use of hard labels, however, is failing to include semantically rich information in the prototypes as well as ignoring the subtle differences in expression between subjects. Recent developments in Vision-Language Models (VLM)~\cite{radford_CLIP_2021, jia2021align, alayrac2022flamingo}, using image-caption pairs instead of hard-labels have demonstrated superior generalisation and zero-shot abilities. Furthermore, CLIP~\cite{radford_CLIP_2021} has been used as the basis for several methods in action recognition~\cite{zara_autolabel_2023, wang_actionclip_2021} or video retrieval and captioning~\cite{tang_clip4caption_2021, ma_x-clip_2022, yang_vid2seq_2023, luo_clip4clip_2022}. However, the use of VLMs in dynamic FER remains relatively unexplored. Recent preliminary works in FER have used video-language contrastive training~\cite{li_cliper_2023, li_fer-former_2023}, primarily relying on class-level prompt learning. Nonetheless, this approach is akin to class supervision and is not designed to generalise to unseen behaviours. 

In this work, we propose a novel approach to zero-shot FER from video inputs by jointly learning video and text embeddings, utilising the contrastive learning framework with natural language supervision. The network architecture is simple and trains a video and text encoder concurrently, as shown in Fig.~\ref{fig:overview}.  The text and image encoders are initialised by leveraging the knowledge of large-scale pre-trained CLIP~\cite{radford_CLIP_2021}, as typical dynamic FER datasets do not have enough samples to train from scratch, and we train an additional temporal module from scratch, similar to previous works in action recognition~\cite{lin_frozen_2022, wang_actionclip_2021}.
Contrary to previous VLM or ZSL works in emotion recognition~\cite{banerjee_learning_2022, zhan_zero-shot_2019, li_cliper_2023, li_fer-former_2023} we use sample-level descriptions during training i.e. captions of the subject's facial expression and context, available in the MAFW~\cite{liu_mafw_2022} dataset. These act as soft labels and aim to achieve more semantically rich latent representations of the video samples. Then, during inference, we use class-level descriptions for each emotion. Specifically, we generate descriptions of each emotion in relation to the typical facial expressions associated with it. In the case of compound emotions, we propose manipulating the latent representation of the categories' descriptions in the embedding space rather than creating additional prompts. Specifically, as compound emotions are combinations of basic emotions, we propose averaging the embeddings of the components and adding them to the set of embeddings for each additional compound emotion.  We show that the proposed methodology trained on sample-level descriptions shows generalisation capabilities invariant to domain shift as we perform zero-shot evaluation on multiple datasets. We also show that compared to the CLIP and FaRL baselines, the temporal information and domain-specific knowledge of the sample-level descriptions improve the zero-shot performance of FER. Finally, to show the generalisation ability of the representations we obtain from the video encoder, we adapt them to the domain of mental health. Using a simple MLP, trained in a fully supervised manner, we achieve results comparable to or outperforming previous state-of-the-art on estimating non-verbal symptoms of schizophrenia. We see a significant improvement compared to previous works, particularly on symptoms associated with affect and expression of emotions as well as total negative score, which is similar to human experts.

Our main contributions can be summarised as follows:
\begin{itemize}
    \item Our paper introduces a novel zero-shot Facial Emotion Recognition (FER) paradigm from video input, employing sample-level descriptions and a dynamic model. This straightforward approach, which leverages CLIP~\cite{radford_CLIP_2021}, outperforms class-level descriptions and significantly improves zero-shot classification performance, particularly for under-represented emotions.
    \item We propose a novel method for representing compound emotions using average latent representations of basic emotions instead of concatenating or generating new prompts. This approach is more intuitive and efficient than prompt engineering and shows significant improvements across all metrics compared to prompt concatenation.
    \item Our proposed method, EmoCLIP, trained on the MAFW~\cite{liu_mafw_2022} dataset and evaluated on popular video FER datasets (AFEW~\cite{afew}, DFEW~\cite{dfew}, FERV39K~\cite{ferv39k}, and MAFW), achieves state-of-the-art performance on ZSL. Additionally, we evaluate the embeddings of the video encoder of EmoCLIP on the downstream task of schizophrenia symptom estimation using the NESS dataset~\cite{ness}. We achieve results comparable to or better than previous state-of-the-art methods and comparable to human experts with a simple 2-layer MLP.
\end{itemize}
The remainder of this paper is organised as follows. Section~\ref{sec:related} discusses related literature, Section~\ref{sec:method} introduces our methodology, Section~\ref{sec:res} reports the results on the tasks of zero-shot FER for simple and compound emotions, and the downstream task of schizophrenia symptom estimation, and Section~\ref{sec:conclusion} concludes the paper.

%%%%%%%%%%%%%%%%%%%%%%%%%%%%%%%%%%%%%%%%%%%%%%%%%%%%%%%%%%%%%%%%%%%%%%%%%%%%%%%%%%
%                             Related Work 
%%%%%%%%%%%%%%%%%%%%%%%%%%%%%%%%%%%%%%%%%%%%%%%%%%%%%%%%%%%%%%%%%%%%%%%%%%%%%%%%%%
\section{Related Work}
\label{sec:related}
In this section, we review previous works on Vision and Language Models (VLM) and zero-shot learning in FER and mental health.

\subsection{Vision and Language Models}

The use of large contrastive pre-training for Vision-Language Models (VLM) has become increasingly popular, as these models have demonstrated impressive generalisation capabilities~\cite{radford_CLIP_2021, jia2021align, zheng_FARL_2022, alayrac2022flamingo}. As large VLM require a large number of data and very high computational resources to achieve those results, the latest research on VLM has been mostly concentrated on three paths: (a) latent space manipulation, (b) leveraging pre-trained spatial features and fine-tuning a temporal module for video input, and (c) prompt learning.  

Menon \& Vondrick~\cite{menon2022visual} use an ensemble of prompts generated by Large Language Models (LLMs) containing descriptive features of each class and show significant improvements in terms of accuracy as well as explainability of decisions. Ouali~\etal~\cite{ouali_black_2023} propose a method for latent space feature alignment in a target domain without the need for additional training. Similarly, Bain~\etal~\cite{bain_clip-hitchhikers_2022} propose several methods for temporal pooling of frames, using pre-trained VLMs with very little or no additional training.
Such approaches, although effective, use no information from the temporal dimension in the video, which is essential in human FER to understand macro and micro-expressions. 

Large VLMs trained on static images have been used for video classification, particularly for action recognition. Lin~\etal~\cite{lin_frozen_2022} propose using a lightweight Transformer Decoder over the CLIP~\cite{radford_CLIP_2021} spatial features for downstream classification. Similarly, ActionCLIP~\cite{wang_actionclip_2021} class labels are used for natural language supervision; therefore, the open vocabulary capabilities are lost in both approaches. CLIP~\cite{radford_CLIP_2021} has been used as a backbone in several video captioning works~\cite{luo_clip4clip_2022, xue_clip-vip_2022, ma_x-clip_2022}. However, none of these works has been evaluated or trained on the domain of FER, where the behaviour is generally not as clearly defined.

To overcome the challenges of prompt engineering in VLMs, some works propose learning a set of tokens~\cite{zhou_learning_2022, zhou_conditional_2022, parisot_learning_2023} to append to the class name, which can improve performance as the text-encoder acts more like a bag of words~\cite{yuksekgonul_when_2023, bagad_test_2023}.  Natural language supervision for facial expression recognition (FER) is a relatively unexplored idea, but there have been some preliminary works exploring this approach. For instance, CLIPER~\cite{li_cliper_2023} has proposed prompt learning to improve closed dictionary FER. However, these tokens are class-specific and cannot be used in open dictionary settings for zero-shot classification.

\subsection{Zero-Shot Learning in Facial Expression Recognition}

Several works~\cite{banerjee_learning_2022, qi_zero-shot_2021, xu_exploring_2022, xu2023zero} have proposed zero-shot frameworks for emotion recognition or emotional response recognition~\cite{zhan_zero-shot_2019} by aligning class name embeddings to multi-media embeddings and then evaluating the method on unseen emotions. However, these methods still rely on hard labels and simpler class prototype embeddings (such as word2vec). As such, they do not take into consideration the intra-class differences or the underlying concepts in each class and as such, do not capture semantically rich information in the latent representations.

\begin{figure*}[t!]
    \centering
    \includegraphics[width=\textwidth]{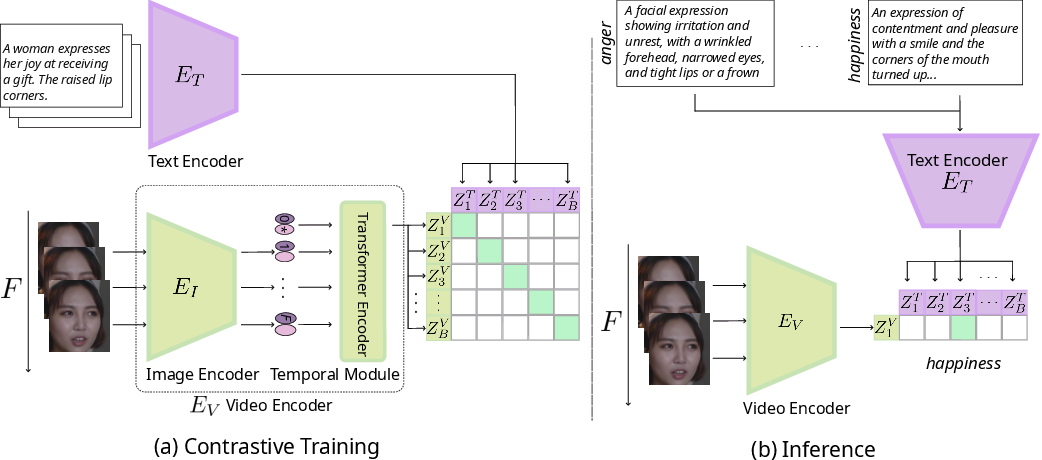}
    \caption{Overview of our method, EmoCLIP. During training (a), we use joint training to optimise the cosine similarity of video-text embedding pairs in the mini-batch. Sample-specific descriptions of the subject's facial expressions are used to train the model. During inference (b), we perform zero-shot classification using class-level descriptions for each of the emotion categories.}
    \label{fig:overview}
\end{figure*}
%%%%%%%%%%%%%%%%%%%%%%%%%%%%%%%%%%%%%%%%%%%%%%%%%%%%%%%%%%%%%%%%%%%%%%%%%%%%%%%%%%
%                             Methodology 
%%%%%%%%%%%%%%%%%%%%%%%%%%%%%%%%%%%%%%%%%%%%%%%%%%%%%%%%%%%%%%%%%%%%%%%%%%%%%%%%%%
\section{Methodology}
\label{sec:method}
% \subsection{Overview}

An overview of the proposed method can be seen in Fig.~\ref{fig:overview}. In a nutshell, we follow the  CLIP~\cite{radford_CLIP_2021}  contrastive training paradigm to optimise a video and a text encoder jointly. The video and text encoders of the network are jointly trained using a contrastive loss over the cosine similarities of the video-text pairings in the mini-batch. 
More specifically, the video encoder ($E_V$) is composed of the CLIP image encoder ($E_I$) and a Transformer Encoder to learn the temporal relationships of the frame spatial representations. The text encoder ($E_T$) used in our approach is the CLIP text encoder. The weights of the image and text encoders in our model are initialised using the large pre-trained weights of CLIP~\cite{radford_CLIP_2021} and finetuned on the target domain, as FER datasets are not large enough to train a VLM from scratch with adequate generalisation. Contrary to the previous video, VLM works in both action recognition~\cite{wang_actionclip_2021, lin_frozen_2022} and FER~\cite{li_cliper_2023, li_fer-former_2023}, we propose using sample level descriptions for better representation learning, rather than embeddings of class prototypes. This leads to more semantically rich representations, which in turn allows for better generalisation. 

We describe how we train the proposed method for dynamic FER in Section~\ref{subsec:proposed}, and how we use it at inference time for simple and compound emotions in Section~\ref{sec:compound}

\subsection{Architecture and Training}
\label{subsec:proposed}
The CLIP framework, proposed by Radford~\etal~\cite{radford_CLIP_2021}, operates as a contrastive multi-modal system, encoding image and text features into a shared space and maximising the cosine similarity of matching image-text pairs (positives) while minimising the similarity of all other pairs (negatives) by optimising a cross-entropy loss over the similarity pairs. We adopt the framework in the dynamic paradigm by introducing a transformer encoder over the spatial features to learn from the temporal dimension.

More specifically, we utilise two separate encoders for processing video and text inputs. Given a video-text pair $x = \{x^V, x^T\}$, we obtain the video-text embeddings using the respective encoders so that $\textbf{z}^V = E_V(x^V)$ and $\textbf{z}^T = E_T(x^T)$, where $\textbf{z}^V, \textbf{z}^T \in \mathbb{R}^D$. The video and text embeddings, $\textbf{z}^V$ and $\textbf{z}^T$, respectively, obtained for each video-text pair in the mini-batch $B$, are utilised to generate a $B \times B$ matrix of cosine similarities. The diagonal elements of the matrix correspond to the $B$ positive pairings, while the remaining elements represent $B^2 - B$ negative pairings. A cross-entropy loss is employed to maximise the similarity between the positives on the diagonal and minimise the similarity of the negatives. 

The video encoder architecture is relatively simple and similar to architectures proposed for video captioning~\cite{ma_x-clip_2022, tang_clip4caption_2021}. Specifically, ($E_V$) is composed of the CLIP image encoder ($E_I$) that extracts frame-level features, which are then fed to a two-layer transformer encoder that acts as a temporal encoder. The state of the learnable classification token at the output of the transformer is used as the video embedding $\textbf{z}^V$. 

While the encoders $E_V, E_T$ could be trained from scratch given a sufficiently large dataset of video-caption pairs, in the domain of FER, the only available dataset with such annotations that do not explicitly include emotional categories in the descriptions is the MAFW~\cite{liu_mafw_2022} dataset, which is relatively small. To this end, we leverage the pre-trained CLIP~\cite{radford_CLIP_2021} image and text encoders to initialise the weights of $E_I$ and $E_T$ in our architecture and fine-tune on the target domain.

\subsection{Inference}
\label{sec:inference}

During inference, the cosine similarity of text and image embedding in the joined latent space is used as the basis for the classification, as described by Radford~\etal~\cite{radford_CLIP_2021}. The prediction probability is then defined as:
\begin{equation}
    P(y = i | x) = \frac{e^{\langle \textbf{z}^V, \textbf{z}^T_i \rangle / \tau}}{\sum_{j=1}^N e^{\langle \textbf{z}^V, \textbf{z}^T_j \rangle / \tau}},
\end{equation}
where $\tau$ is a learnable temperature parameter in CLIP and $\langle \cdot, \cdot \rangle$ is the cosine similarity. 

\subsection{Class Descriptions}

In the case of basic emotions, we provide class descriptions in the form of natural language obtained from LLMs, rather than using a prompt in the form of \textit{`an expression of \{emotion\}'}, to match the information-rich sample level descriptions. The use of descriptions is intuitive in the context of emotions and FER, as using the class names would imply that the model has a deep understanding of emotional expression (e.g. in Fig.~\ref{fig:overview} ``raised lip corners'' and ``happiness'' are not directly associated). In addition, due to the very large intra-class variation of emotional expression, the definitions of emotions are fluid. Therefore, descriptions should be more fitted to differentiate fine-grained emotions.
We note that these are different descriptions than the ones used during training, as in the latter, we use sample-level video-text pairs, as can be seen in Fig.~\ref{fig:overview}. As such, our method is performing zero-shot classification during inference.
Specifically, to obtain the class-level descriptions, we prompt ChatGPT with the input:

\textit{Q: What are the facial expressions associated with \{emotion\}?}

We then curate the generated responses to exclude irrelevant information, such as body pose or emotional cues (such as ``a sad expression'' for helplessness). For example, \textit{``A facial expression showing irritation and unrest, with a wrinkled forehead, narrowed eyes, and tight lips or a frown''} is the generated description for \textit{``anger''}. Examples of prompts can be seen in Fig.~\ref{fig:overview} \&~\ref{fig:compound}; the full set of class descriptions used for inference are given in the Supplementary Material. This strategy is proposed over the prompt templates used in CLIP~\cite{radford_CLIP_2021}, as the prompt in the form of \textit{`an expression of \{emotion\}'} would imply a universal definition for each emotion, that the text encoder has learnt. Additionally, while the CLIP prompts have shown impressive results on clearly defined objects in images, emotions are significantly more vague and open to interpretation, both in terms of expression and understanding. The MAFW~\cite{liu_mafw_2022} dataset does not include sample-level descriptions for the neutral category; as such, we generate descriptions for neutral samples by randomly selecting and concatenating two prompts generated from ChatGPT for the neutral category. The full list of prompts for the neutral category is given in the Supplementary Material.
The use of LLM, in this case, ChatGPT, over hand-crafted prompts is proposed to avoid introducing the authors' bias in the prompts.

\label{sec:compound}
\begin{figure}
  \begin{center}
    \includegraphics[width=0.5\textwidth]{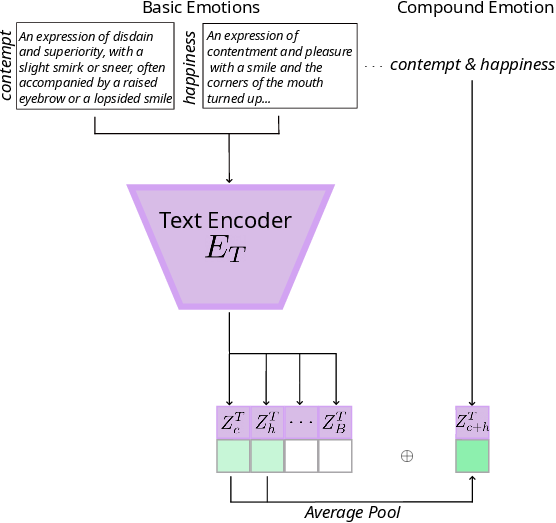}
  \end{center}
    \caption{EmoCLIP manipulates the latent space of basic emotions to create representations for compound emotions. We take the average latent representation of the components and concatenate it to the set of representations for each new compound emotion.}
  \label{fig:compound}
\end{figure}

Compound emotions are a complex combination of basic emotions, such as ``happily surprised''~\cite{cowen_self-report_2017, liu_mafw_2022}, that are used to identify a wider range of human facial expressions. As, by definition, compound emotions are combinations of basic emotions, we propose a new approach to constructing the latent $\textbf{z}^T$ representation. Instead of treating them as independent emotional states and creating additional prompts, we use the pre-normalised latent representations of the components to compose the new compound emotion as shown in Fig.~\ref{fig:compound}. 

Formally, for any new compound emotion, we calculate its latent representation $\textbf{z}_n^T$ as the average of the latent representations of its $C$ component emotions so that: 
\begin{equation}
 \textbf{z}_n^T = \frac{1}{C} \sum_{c=0}^C \textbf{z}_c^T    
\end{equation}
The resulting vector representations are then concatenated to the set of class representations, and inference is performed over the latent representations of basic and compound emotions.

%%%%%%%%%%%%%%%%%%%%%%%%%%%%%%%%%%%%%%%%%%%%%%%%%%%%%%%%%%%%%%%%%%%%%%%%%%%%%%%%%%
%                             Experimental Results 
%%%%%%%%%%%%%%%%%%%%%%%%%%%%%%%%%%%%%%%%%%%%%%%%%%%%%%%%%%%%%%%%%%%%%%%%%%%%%%%%%%
 
\section{Experimental Results}
\label{sec:res}

In this section, we present the experimental set-up (Section~\ref{sec:setup}), the ablation study (Section~\ref{sec:ablation}), and the experimental evaluation of the proposed framework in the Zero-Shot paradigm (Section~\ref{sec:zero}) as well as the Downstream task of schizophrenia symptom estimation (Section~\ref{sec:schi}).

\begin{table*}[t!]
% \scriptsize
\centering
\begin{tabular}{lcccl|cc}
\textbf{Architecture}      & \textbf{Temporal Layer}          & \textbf{Prompt Ensemble} & \textbf{Frame Ensemble} & \textbf{Pre-training} & \textbf{UAR} & \textbf{WAR} \\ \hline
CLIP~\cite{radford_CLIP_2021} (Frozen backbone)    &                &                 &                & Laion-400m  & 16.97                            & 21.69                   \\
CLIP~\cite{radford_CLIP_2021} (Frozen backbone)   &                &   \checkmark    &                & Laion-400m  & 19.77                            & 18.64                   \\
CLIP~\cite{radford_CLIP_2021} (Frozen backbone)   &                &   \checkmark    &    \checkmark  & Laion-400m  & 19.46                            & 17.61                   \\
CLIP~\cite{radford_CLIP_2021} (Frozen backbone)   &                &                 &    \checkmark  & Laion-400m  & 20.40                            & 21.16                   \\
CLIP~\cite{radford_CLIP_2021}               &                &                 &    \checkmark  & MAFW       & 17.62                            & 20.57              \\
EmoCLIP (Frozen backbone) & \checkmark    &                 &                & MAFW        & \uline{23.60}            & \uline{31.36} \\
EmoCLIP &  \checkmark    &                 &                & MAFW        & \textbf{25.86}                   & \textbf{33.49}         
\end{tabular}
\caption{Performance of the proposed method, EmoCLIP, on the MAFW~\cite{liu_mafw_2022} dataset on 11-class single expression classification against the baseline, with different frame aggregation and prompting strategies.
}
\label{tab:ablation}
\end{table*}

\subsection{Experimental Setup}
\label{sec:setup}
% \subsubsection{Datasets}
\textbf{Datasets:} We train on the \textbf{MAFW}~\cite{liu_mafw_2022} dataset, as to our knowledge, it is the only FER dataset that includes sample-level descriptions of the subject's context and facial expressions without explicitly mentioning the emotional state. The dataset contains $10k$ audio-video clips with categorical annotations for 11 emotions, accompanied by short descriptions of the subject's facial expressions in two languages, English and Mandarin Chinese; we use the former for all experiments in this work. The dataset is divided into five folds for evaluation.

We evaluate our method on three additional FER datasets on the seven basic emotions. The \textbf{AFEW}~\cite{afew} dataset contains 1,809 clips from movies, divided into three subsets. We train the method from scratch and report results on the validation set, as the labels of the test set are not publicly available. \textbf{DFEW}~\cite{dfew} is composed of 16,000 videos from movies, split into five folds for cross-validation. \textbf{FERV39K}~\cite{ferv39k} is a large dataset with around 39,000 clips annotated by 30 annotators, and we report results on the test set.

Finally, we evaluate the video encoder of EmoCLIP as trained on the MAFW dataset, on the downstream task of schizophrenia symptom estimation using a subset of the \textbf{NESS}\cite{ness} dataset, as described in~\cite{bishay2019schinet, foteinopoulou_learning_2022}. The subset includes 113 in-the-wild baseline clinical interviews from 69 patients and two symptom scales, PANSS\cite{kay_positive_1987} and CAINS\cite{forbes_initial_2010}. To ensure a fair comparison with previous works, we used leave-one-patient-out cross-validation. The values of ``Total Negative'' and ``EXP - Total'' in PANSS and CAINS scales respectively, were scaled during training to match the range of individual symptoms. As the videos in the dataset are very long, ranging between 40 and 120 minutes long, we average the clip-level predictions to obtain the final video-level prediction vector for each video and calculate the network's performance over the video-level predictions during inference, as described in~\cite{foteinopoulou_learning_2022}.

% \subsubsection{Training Details}
\textbf{Training Details:} The CLIP image encoders used in all experiments have a ViT-B/32 architecture. During training, we apply augmentation to the spatial dimensions of the video inputs for all datasets. Specifically, we apply a random horizontal flip to all frames in a sequence with a probability of 50\%. We also apply a random rotation with a range of 6\textdegree and random centre crop. 
In the temporal dimension, we empirically trim all videos to $T=32$ number of frames and downsample by a factor of 4 (which results in the entire video being included for the majority of samples). In the downstream task, the random crop along the temporal dimension acts as an augmentation technique.

For all experiments, we use a Stochastic Gradient Descent optimiser (SGD) with a learning rate of $10^{-3}$. We finetune the pre-trained parameters of the CLIP backbone using a different learning rate of $10^{-6}$ for the image and text encoders.

\subsection{Ablation Study}
\label{sec:ablation}

\begin{table*}[h]
\centering
\begin{tabular}{lll||ll}
Mode                        & \multicolumn{1}{c}{Architecture} & \multicolumn{1}{c||}{Contrastive Pre-training} & \multicolumn{1}{c}{UAR} & \multicolumn{1}{c}{WAR}  \\ 
\hhline{===::==}
\multirow{6}{*}{Supervised} & C3D~\cite{liu_mafw_2022}                              & -                                              & 31.17                   & 42.25                    \\
                            & Resnet18\_LSTM~\cite{liu_mafw_2022}                   & -                                              & 28.08                   & 39.38                    \\
                            & VIT\_LSTM~\cite{liu_mafw_2022}                        & -                                              & 32.67                   & 45.56           \\
                            & C3D\_LSTM~\cite{liu_mafw_2022}                        & -                                              & 29.75                   & 43.76                    \\
                            & T-ESFL~\cite{liu_mafw_2022} & - & 33.28 & \textbf{48.18} \\
                            & EmoCLIP (LP)                & - &           30.26     &       44.231     \\
                            & EmoCLIP (Frozen backbone)           & MAFW [class descriptions]                         & \textbf{34.24}          & 41.46                    \\ 
\hline
\multirow{4}{*}{Zero-shot}  & CLIP~\cite{radford_CLIP_2021}                            & Laion-400m                                     & 20.40                    & 21.16.                       \\
                            & FaRL - ViT-B/16~\cite{zheng_FARL_2022}                            & Laion Face-20M        & 14.07   & 7.70                    \\
                            & EmoCLIP                             & MAFW [sample descriptions]                     & \textbf{25.86}          & \textbf{33.49}          
\end{tabular}
\caption{Performance of the proposed method on the MAFW~\cite{liu_mafw_2022} dataset on 11-class single expression classification against other SOTA architectures in a supervised and zero-shot setting.}
\label{tbl:mafw_zero}
\end{table*}

In order to examine the effect of different components of EmoCLIP and evaluate the prompting strategy used in this work, we perform several experiments using baseline CLIP and our proposed method, EmoCLIP. More specifically, as discussed in the Methodology Section~\ref{sec:method}, during inference, we use class descriptions generated with the help of ChatGPT; this is different to the prompt ensemble of Radford~\etal~\cite{radford_CLIP_2021} who use several prompt templates in the form of \textit{`an expression of \{emotion\}'} and average their latent representation vectors during inference. Furthermore, as CLIP~\cite{radford_CLIP_2021} is a static model, we conduct two zero-shot evaluation strategies, first on the middle frame and then by averaging the latent representations of all frames, i.e. performing frame ensemble. To further show the necessity for a temporal layer, we finetune a CLIP~\cite{radford_CLIP_2021} architecture on the MAFW~\cite{liu_mafw_2022} dataset using frame ensembling, which has a negative effect on CLIP's performance on the task. We theorise that frame ensembling in FER tasks negatively affects training as averaging out frame representations in most cases will consider noisy and keyframes equally, thus muddling the video latent features~\cite{li2022nr}, which confuses the network.

The proposed architecture, which incorporates a temporal layer, has significant performance improvements compared to the baseline CLIP architectures.
Finally, we compare the performance of EmoCLIP with a frozen backbone to that of our proposed method. The results of the ablation study can be seen in Tab.~\ref{tab:ablation}. As the prompt templates used in CLIP~\cite{radford_CLIP_2021} are similar to the format ``A photo of {class name}'', they are more suitable for static images rather than video. This is corroborated by the performance of the CLIP~\cite{radford_CLIP_2021} baseline using prompt ensemble vs our class descriptions. Additionally, as emotional expression is not static, we also observe an increase in the baseline performance using frame ensembling compared to evaluating on the middle frame. Such an outcome is intuitive as dynamic FER has a wider temporal context that needs to be considered rather than a single frame. Furthermore, the temporal relationships between frames hold important information in FER tasks.
This is also confirmed by the proposed architecture, including a temporal module, which offers a large increase in performance for both Weighted Average Recall (WAR) and Unweighted Average Recall (UAR). By finetuning the backbone to the FER domain, we see a further increase in performance by approximately 2\% for each metric. 

\begin{table}[h]
\footnotesize
% \scriptsize
\setlength{\tabcolsep}{2pt}
\centering

\begin{tabular}{llcllll}
Mode                        & \multicolumn{1}{c}{Architecture} & Repr. Avg & \multicolumn{1}{c}{UAR} & \multicolumn{1}{c}{WAR} & F1            & AUC             \\ 
\hline\hline
\multirow{5}{*}{Supervised} & C3D~\cite{liu_mafw_2022}                              & -                                              & \textbf{9.51}           & 28.12                   & 6.73          & 74.54           \\
                            & Resnet18\_LSTM~\cite{liu_mafw_2022}                  & -                                              & 6.93                    & 26.6                    & 5.56          & 68.86           \\
                            & VIT\_LSTM~\cite{liu_mafw_2022}                        & -                                              & 8.72                    & 32.24                   & \textbf{7.59} & 75.33           \\
                            & C3D\_LSTM~\cite{liu_mafw_2022}                        & -                                              & 7.34                    & 28.19                   & 5.67          & 65.65           \\
                            & T-ESFL~\cite{liu_mafw_2022}                           & -                                              & 9.15                    & \textbf{34.35}          & 7.18          & \textbf{75.63}           \\ 
\hline
\multirow{5}{*}{Zero-shot}  & Random                             &  -   & 2.38                   & 7.72                    & 0.34          & 50.00           \\
                            & CLIP~\cite{radford_CLIP_2021}     &  \xmark   & 4.72                    & 5.25                    & 2.44          & 51.89           \\
                            & CLIP~\cite{radford_CLIP_2021}     & \checkmark & 4.14                             & 5.35                             & 2.46          & \textbf{53.07} \\
                            & FaRL~\cite{zheng_FARL_2022}                             &  \xmark   & 3.03                    & 4.66                    & 2.16          & 51.01           \\
                            & FaRL~\cite{zheng_FARL_2022}      & \checkmark & 4.00                             & 5.75                             & 2.56          & 51.10          \\
                            & EmoCLIP                          &  \xmark         & 5.24                    & 15.34                   & 3.80          & 51.30           \\
                            & EmoCLIP                          &   \checkmark     & \textbf{6.58}           & \textbf{18.53}          & \textbf{4.78} & \u{52.59} 
\end{tabular}
\caption{Zero-shot classification on the 43 compound expressions of the MAFW~\cite{liu_mafw_2022} dataset. Supervised methods are included as a reference.}
\label{tbl:compound}
\end{table}

\subsection{Zero-Shot Evaluation}
\label{sec:zero}
To evaluate the effectiveness of the proposed method, we compare it with pre-trained CLIP~\cite{radford_CLIP_2021} and FaRL~\cite{zheng_FARL_2022} models in a Zero-shot setting. As both of these methodologies are trained on static images, we take the average of the latent representations of all frames in a video to compute the video embedding and use that to calculate the cosine similarity with the text description embeddings. 
We show the performance of our method against the CLIP and FaRL baselines, with a frozen CLIP backbone and the finetuned image-text encoders on the 11 class classification of MAFW~\cite{liu_mafw_2022} in Table~\ref{tbl:mafw_zero}. We note that even though FaRL~\cite{zheng_FARL_2022} is trained on a subset of the Laion dataset~\cite{laion} filtered to include samples of faces, the model trained on FaRL performs significantly worse than both the CLIP baseline and our proposed method. We also note that the FaRL pre-trained weights are only available for the ViT-B/16 architecture, which may explain the difference in performance. Furthermore, while FaRL is trained with image-text pairs of faces, these are not necessarily related to facial expression or are not well grounded with text, as we can see by visually inspecting the LAION-face dataset~\cite{zheng2022general}, so we hypothesise that the FaRL embedding space is significantly more niche than CLIP but not in a direction beneficial for zero-shot FER. 

\begin{figure*}[t]
% \begin{tabular}{c}
\centering
\includegraphics[width=\textwidth]{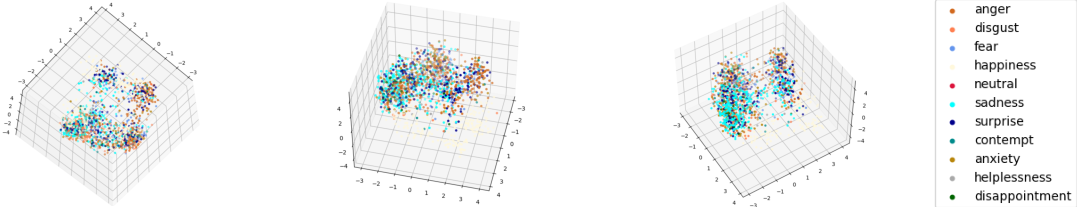} \\
\centering
(a) CLIP Latent Image Representations \\
\centering
\includegraphics[width=\textwidth]{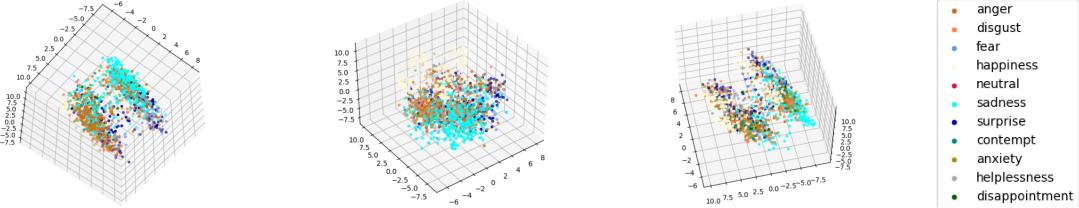} \\
(b) EmoCLIP Latent Image Representations
% \end{tabular}
\caption{Latent representation of the MAFW dataset using CLIP (a) and EmoCLIP (b) for each emotion category.}
\label{fig:subspaces}
\end{figure*}

\begin{table*}[h]
\footnotesize
\centering
\begin{tabular}{ll|l||ll|ll|ll|ll}
                            &                               &                       & \multicolumn{2}{l|}{\begin{tabular}[c]{@{}l@{}}DFEW\\(7 classes)\end{tabular}} & \multicolumn{2}{l|}{\begin{tabular}[c]{@{}l@{}}AFEW\\(7 classes)\end{tabular}} & \multicolumn{2}{c|}{\begin{tabular}[c]{@{}c@{}}FERV39K\\(7 classes)\end{tabular}} & \multicolumn{2}{c}{\begin{tabular}[c]{@{}c@{}}MAFW\\(11 classes)\end{tabular}}                  \\
                            & Architecture                  & Training labels       & UAR            & WAR                                                           & UAR            & WAR                                                           & UAR            & WAR                                                              & UAR                                       & WAR                                                 \\ 
\hhline{===::========}
\multirow{2}{*}{Supervised} & EmoCLIP                       & {[}class description] & \textbf{58.04} & \textbf{62.12}                                                & \textbf{44.32} & \textbf{46.19}                                                & \textbf{31.41} & 36.18                                                            & \textbf{34.24}                            & 41.46                                               \\
                            & EmoCLIP (LP)                  & {[}class]             & 50.29          & 62.09                                                         & 33.74          & 38.85                                                         & 30.58          & \textbf{43.54}                                                   & \textcolor[rgb]{0.102,0.102,0.102}{30.26} & \textcolor[rgb]{0.102,0.102,0.102}{\textbf{44.21}}  \\ 
\hhline{===::========}
\multirow{3}{*}{Zero-shot}  & CLIP~\cite{radford_CLIP_2021}  & {[}image caption]     & 19.86          & 10.60                                                         & 23.05          & 11.80                                                         & 20.99          & 17.10                                                            & 20.04                                     & 21                                                  \\
                            & EmoCLIP (leave-one-class-out) & {[}class description] & \uline{22.85}  & \uline{24.96}                                                 & \uline{35.11}  & \uline{27.57}                                                 & \textbf{39.35} & \textbf{41.60}                                                   & \uline{24.12}                             & \uline{24.74}                                       \\
                            & EmoCLIP                       & {[}video caption]     & \textbf{36.76} & \textbf{46.27}                                                & \textbf{36.13} & \textbf{39.90}                                                & \uline{26.73}  & \uline{35.30}                                                    & \textbf{25.86}                            & \textbf{33.49}                                     
\end{tabular}
\caption{Evaluation of EmoCLIP using sample descriptions vs class-level description as natural language supervision, on four video FER datasets.}
\label{tbl:supervised}
\end{table*}

To further investigate the improvement of our method vs baseline CLIP, we use PCA to reduce the high-dimensional latent image vectors to three dimensions (which explain approximately 70\% of the variance) and plot them in a 3D scatter plot, as shown in Fig.~\ref{fig:subspaces}. We see that by fine-tuning to the FER domain, the categorical emotions form more distinct clusters than in CLIP, which is also reflected in the classification performance of our method. The more discriminate latent space is somewhat expected as the CLIP domain is much larger; thus, emotions and FER occupy a much smaller subspace, i.e. emotional categories will be closer to each other than to animals. By fine-tuning, the distances between dissimilar samples become larger, which allows for more discrete clusters.

For reference, we also train our architecture with class-level descriptions, and using an MLP head with two fully connected layers (Linear probe shown as LP on the table) over the EmoCLIP video encoder, we show the performance against several supervised architectures as reported in~\cite{liu_mafw_2022}. We see that the architecture trained using the class descriptions outperforms previous methods in terms of UAR and is comparable with others in terms of WAR. These results indicate that the contrastive vision-language approach leads to more semantically rich and discriminate latent representations even in a supervised setting. The difference in the two metrics is somewhat expected, as FER datasets are typically imbalanced.

Furthermore, we present experimental results on the classification of 43 compound emotions in the MAFW dataset in Table~\ref{tbl:compound}. We evaluate the performance of our proposed method, EmoCLIP, against a baseline approach of using concatenated prompts, as well as CLIP and FaRL baselines. Specifically, we concatenate the class descriptions for each compound emotion and use this as class prompt input. We demonstrate that EmoCLIP outperforms the baseline approach for all metrics. Moreover, we note that in the 43 emotions classification, both CLIP and FaRL perform significantly worse than EmoCLIP and have performance comparable to random (where only the majority class is predicted). We theorise this is due to the lack of temporal understanding of the static models. Furthermore, without fine-tuning on the target domain, the class descriptions are not discriminate enough for the zero-shot classification of emotions, as previously discussed. However, the representation average method on both baselines improves the performance of the static models on the compound emotions task. We also report the results of several supervised methods for reference, which perform significantly better than zero-shot approaches as expected. 

Finally, we evaluate the performance of our proposed method using sample-level descriptions from MAFW~\cite{liu_mafw_2022} on four widely used video FER datasets and compare it with the CLIP baseline as shown in Table~\ref{tbl:supervised}. Additionally, in line with previous works in zero-shot emotion classification~\cite{banerjee_learning_2022, xu_exploring_2022, zhan_zero-shot_2019, qi_zero-shot_2021}, we train our architecture using class-level descriptions and evaluate using leave-one-class-out (loco) cross-validation. We note that we cannot directly compare with these architectures, as they involve either different modalities (e.g. audio, pose)~\cite{banerjee_learning_2022, xu_exploring_2022, qi_zero-shot_2021, xu2023zero} or a different task~\cite{zhan_zero-shot_2019}, we adopt however, their experimental set-up using our architecture to show how natural language supervision and semantically rich class descriptions can help improve zero-shot FER performance.

We observe that the EmoCLIP trained on MAFW~\cite{liu_mafw_2022} sample-level descriptions show impressive generalisation ability on all datasets that we evaluate. Specifically, for AFEW~\cite{afew}, MAFW~\cite{liu_mafw_2022} and DFEW~\cite{dfew}, we see that the EmoCLIP model is outperforming both the loco experiment and the CLIP~\cite{radford_CLIP_2021} baseline. Furthermore, the generalisation of the method is resistant to domain shift from unseen datasets, as we observe from the significant performance increase between the CLIP~\cite{radford_CLIP_2021} baseline and EmoCLIP. We note that for FERV39K~\cite{ferv39k}, the loco experiment has a higher performance than the sample-wise training. However, it is very important to stress that the FERV39K~\cite{ferv39k} is significantly larger than the base dataset (over 3x more samples); therefore, methods trained on it would have an advantage, particularly as in the loco experiment, there is no domain shift.

 \begin{table*}[!h]
    % \footnotesize
    % \scriptsize
    % \setlength{\tabcolsep}{2pt}
    \centering
    \begin{tabular}{l||lll||lll||lll||lll}
             & \multicolumn{3}{c||}{N3: Poor Rapport}        & \multicolumn{3}{c||}{N6: Lack of Spont.}     & \multicolumn{3}{c||}{N1: Blunted Affect}     & \multicolumn{3}{c}{Total Negative Score}            \\ 
\cline{2-13}
             & MAE            & RMSE          & PCC          & MAE           & RMSE          & PCC          & MAE           & RMSE          & PCC          & MAE           & RMSE          & PCC           \\ 
\hhline{=::===::===::===::===}
Tron~\etal~\cite{tron_automated_2015}       & 0.98           & 1.31          & .20          & 1.37          & 1.69          & .13          & 0.90          & 1.28          & .37          & -             & -             & -             \\
Tron~\etal~\cite{tron_facial_2016}       & 1.01           & 1.26          & .15          & 1.32          & 1.62          & .09          & 0.99          & 1.36          & .11          & -             & -             & -             \\
SchiNet~\cite{bishay2019schinet}      & \textbf{0.85 } & 1.20          & .27          & 1.25          & 1.51          & .25          & 0.84          & 1.18          & .42          & 3.30          & 4.17          & .29           \\
Relational~\cite{foteinopoulou_learning_2022}  & \uline{0.87}   & \textbf{1.16} & \textbf{.78} & \textbf{0.74} & \textbf{0.95} & \textbf{.47} & \uline{0.64}  & \uline{0.87}  & \uline{.56}  & \uline{2.80}  & \uline{3.78}  & \uline{.71}   \\
EmoCLIP (LP) & 0.89           & 1.28          & \uline{.72}  & \uline{0.91}  & \uline{1.15}  & \uline{.27}  & \textbf{0.56} & \textbf{0.79} & \textbf{.63} & \textbf{2.31} & \textbf{3.01} & \textbf{.85} \\
\end{tabular}
\caption{Performance on the downstream task against other SoTA (PANSS-NEG).}
\label{tbl:panss}
\end{table*}

\begin{table*}[!h]
    \footnotesize
    \setlength{\tabcolsep}{2pt}
    \centering
\begin{tabular}{l||lll||lll||lll||lll||lll}
             & \multicolumn{3}{c||}{Facial Exp.}            & \multicolumn{3}{c||}{Vocal Exp.}             & \multicolumn{3}{c||}{Expr. Gestures}         & \multicolumn{3}{c||}{Quant. of Speech}       & \multicolumn{3}{c}{EXP-Total Score}           \\ 
\cline{2-16}
             & MAE           & RMSE          & PCC          & MAE           & RMSE          & PCC          & MAE           & RMSE          & PCC          & MAE           & RMSE          & PCC          & MAE           & RMSE          & PCC           \\ 
\hhline{=::===::===::===::===::===}
Tron~\etal~\cite{tron_automated_2015}       & 0.80          & 1.03          & .37          & 0.87          & 1.23          & .23          & 0.85          & 1.19          & .36          & 1.09          & 1.43          & .27          & -             & -             & -             \\
Tron~\etal~\cite{tron_facial_2016}       & 0.75          & 1.07          & .36          & 0.86          & 1.22          & .26          & 0.91          & 1.22          & .38          & 1.02          & 1.36          & .25          & -             & -             & -             \\
SchiNet~\cite{bishay2019schinet}     & 0.66          & 0.93          & .46          & 0.77          & 1.10          & .27          & 0.90          & 1.15          & .36          & 0.98          & 1.30          & .30          & 2.67          & 3.34          & .45           \\
Relational~\cite{foteinopoulou_learning_2022}  & 0.56          & 0.72          & .75          & 0.65          & 0.89          & .71          & 0.71          & 0.89          & \textbf{.76} & \textbf{0.60} & \textbf{0.82} & \textbf{.54} & 1.88          & 2.60          & .77           \\
EmoCLIP (LP) & \textbf{0.49} & \textbf{0.65} & \textbf{.77} & \textbf{0.59} & \textbf{0.83} & \textbf{.74} & \textbf{0.66} & \textbf{0.85} & \uline{.74}  & \uline{0.64}  & \uline{0.89}  & \uline{.50}  & \textbf{1.32} & \textbf{2.52} & \textbf{.83} \\
\end{tabular}
\caption{Performance on the downstream task against other SoTA (CAINS-EXP).}
\label{tbl:cains}
\end{table*}

For reference and to provide context, we include the performance of the architecture in a supervised setting on all four datasets, as there are not many methods performing zero-shot classification in FER. The supervised methods outperform zero-shot, as expected. However, we can compare the performance of our proposed architecture trained contrastively with class-level descriptions and using an MLP head for multi-class classification (Linear Probe). We want to point out that the architecture trained using class descriptions is significantly outperforming the linear probe architecture, particularly in terms of UAR, showing again that natural language supervision can provide significant advantages even in a fully supervised setting. This is especially true for under-represented classes, as indicated by the increase in UAR, which is sensitive to the long tail of the distribution.

\subsection{Downstream Task}
\label{sec:schi}

We evaluate the representations obtained by the proposed method on the downstream task of schizophrenia symptom estimation on two scales, namely the PANSS~\cite{kay_positive_1987} and CAINS~\cite{forbes_initial_2010} scales. As symptoms in both scales have ordinal labels, we address the problem of symptom estimation as a multi-label regression.
We evaluate the proposed method in terms of Mean Absolute Error (MAE), Root Mean Squared Error (RMSE) and Pearson's Correlation Coefficient (PCC) for a fair comparison with previous works in the literature.
We train the linear probe architecture, with the video encoder obtained through contrastive pre-training on the MAFW~\cite{liu_mafw_2022} frozen, and updating only the MLP weights.
The results against previous state-of-the-art (SoTA) can be seen on Tables~\ref{tbl:panss} \&~\ref{tbl:cains} for the PANSS~\cite{kay_positive_1987} and CAINS~\cite{forbes_initial_2010} scales respectively. The proposed method, pre-trained contrastively on sample level descriptions, outperforms or is comparable to previous methods on all symptoms, particularly for ``N1: Blunted Affect'' and ``Facial Expression'' on the PANSS and CAINS scales, which are by definition most related to apparent affect and FER.

As the NESS~\cite{ness} dataset is annotated by multiple healthcare professionals, we can compare the PCC achieved by the method to the annotators' PCC which has a mean value of 0.85 across all symptoms~\cite{bishay2019schinet, ness}. The proposed method on the downstream task achieves performance comparable to that of human experts on the total scores of both scales, as we see in Tables~\ref{tbl:panss} \&~\ref{tbl:cains}.

%%%%%%%%%%%%%%%%%%%%%%%%%%%%%%%%%%%%%%%%%%%%%%%%%%%%%%%%%%%%%%%%%%%%%%%%%%%%%%%%%%
%                             Conclusion 
%%%%%%%%%%%%%%%%%%%%%%%%%%%%%%%%%%%%%%%%%%%%%%%%%%%%%%%%%%%%%%%%%%%%%%%%%%%%%%%%%%

\section{Conclusion}
\label{sec:conclusion}

In this work, we presented a novel contrastive pre-training paradigm for video FER, trained on video-text pairs with sample-level descriptions without any class-level information. While contrastive learning and natural language supervision have been used in other domains, zero-shot emotion recognition remains surprisingly unexplored, with works focusing on creating class prototypes with simpler word encoding methods~\cite{banerjee_learning_2022, qi_zero-shot_2021, xu_exploring_2022, xu2023zero}. Emotional prototypes, however, disregard the intra-class variation that is inherently present in FER tasks. To overcome the limitations of training on coarse emotional categories, EmoCLIP is trained on sample-level descriptions.
We evaluate our method on four popular FER video datasets~\cite {afew, dfew, ferv39k, liu_mafw_2022} and test using zero-shot evaluation on the basic emotions as well as compound emotions. Our method outperforms the CLIP baseline by a large margin and shows impressive generalisation ability on unseen datasets and emotions. To our knowledge, this is the first work to train with sample-level descriptions for FER and to propose zero-shot evaluation using semantically rich class descriptions in the domain. We also evaluated the EmoCLIP video encoder features on schizophrenia symptom estimation, outperforming previous state-of-the-art methods and achieving performance comparable to human experts in terms of PCC.

\clearpage

\begin{figure*}[h]
    \centering
    \includegraphics[width=0.9\textwidth]{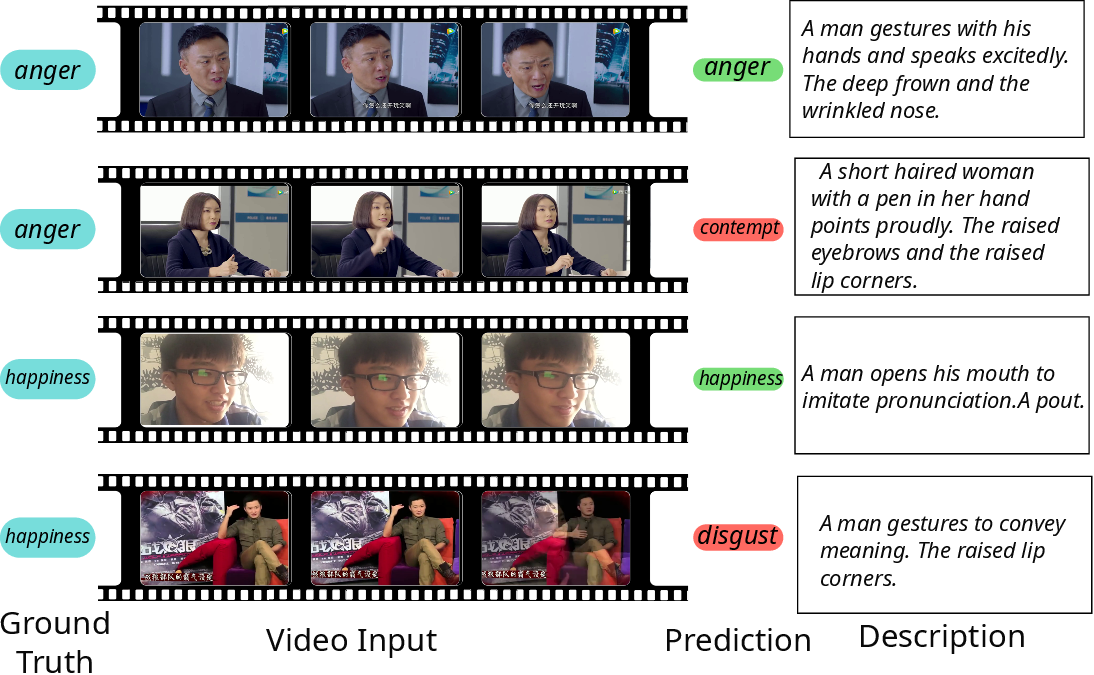}
    \caption{Example frames of correctly and incorrectly classified samples from the MAFW~\cite{liu_mafw_2022} dataset, for anger (top) and happiness (bottom). The sample-level descriptions are included for reference but are not used during inference.}
    \label{fig:quality}
    \vspace{-10pt}
\end{figure*}
\appendix

In this supplementary document, we present additional qualitative results that complement and extend the findings outlined in the main text. More specifically, in~\cref{sec:classdesc}, we show the detailed class descriptions used by our methods. Then, in~\cref{sec:qual}, we show examples of correctly and incorrectly classified samples.

\section*{Class Descriptions}
\label{sec:classdesc}

As described in Section~\ref{sec:compound}, we prompt a popular Large Language Model (LLM), specifically ChatGPT, to obtain the descriptions for each class. The descriptions are then curated manually to exclude information unrelated to the facial expression (e.g. body pose). The final list of descriptions can be seen in Table~\ref{tbl:classdescr}.

\begin{table}[h]
    \footnotesize
    \centering
    \begin{tabular}{cl}
    Class Name     & Description                                                                                                                                                                                                                           \\ 
    \hline\hline
    anger          & \begin{tabular}[l]{@{}l@{}}A facial expression showing irritation and \\ unrest, with a wrinkled forehead, narrowed eyes, \\and tight lips or a frown\end{tabular}                                                                       \\
    disgust        & \begin{tabular}[l]{@{}l@{}}An expression of repulsion and displeasure, with \\ a raised upper lip, a scrunched nose, and a \\downturned mouth\end{tabular}                                                                               \\
    fear           & \begin{tabular}[l]{@{}l@{}}An expression of tension and withdrawal, with \\ wide-open eyes, raised eyebrows, and a slightly \\open mouth. The face may appear physically tense \\ or frozen in fear\end{tabular}                            \\
    happiness      & \begin{tabular}[l]{@{}l@{}}An expression of contentment and pleasure, with a \\ smile and the corners of the mouth turned up, \\often accompanied by crinkling around the eyes. \\ The face may appear relaxed and at ease\end{tabular}     \\
    neutral        & \begin{tabular}[l]{@{}l@{}}An expression of calm and neutrality, with a neutral \\ mouth and no particular indication of emotion. \\The eyebrows are usually not raised or furrowed\end{tabular}                                         \\
    sadness        & \begin{tabular}[l]{@{}l@{}}An expression of sadness and sorrow, with a \\ downturned mouth or frown, and sometimes tears \\or a tightness around the eyes. The face may appear \\ physically withdrawn or resigned\end{tabular}             \\
    surprise       & \begin{tabular}[l]{@{}l@{}}An expression of shock and astonishment, with \\ wide-open eyes and raised eyebrows, sometimes \\accompanied by a gasp or an open mouth\end{tabular}                                                          \\
    contempt       & \begin{tabular}[l]{@{}l@{}}An expression of disdain and superiority, with a slight \\ smirk or sneer, often accompanied by a raised \\eyebrow or a lopsided smile\end{tabular}                                                           \\
    anxiety        & \begin{tabular}[l]{@{}l@{}}An expression of worry and apprehension, with\\ furrowed eyebrows and a tight mouth. \\The eyes may appear wide and darting, and the face \\ may appear physically tense or worried\end{tabular}                \\
    helplessness   & \begin{tabular}[l]{@{}l@{}}An expression of defeat and resignation, with the \\ eyes looking down and the mouth turned down. \\The eyebrows may be furrowed, and the face \\ may appear resigned or resigned\end{tabular}                   \\
    disappointment & \begin{tabular}[l]{@{}l@{}}An expression of frustration and disillusionment, \\ with a slight frown or drooping of the mouth. \\The eyebrows may be lowered or furrowed, and the \\ face may appear physically drawn or tired\end{tabular}  \\
    \hline\hline
    \end{tabular}
    \caption{Class descriptions for each emotion used during inference}
    \label{tbl:classdescr}
\end{table}

The MAFW~\cite{liu_mafw_2022} dataset does not include sample-level descriptions for the neutral category; as such, we use variations of the neutral description randomly for each sample in the neutral category. The list of neutral descriptions for samples with no description during training can be seen in Table.~\ref{tbl:neutral}. 

\begin{table}[h]
\footnotesize
\centering
\begin{tabular}{cl}
\hline\hline
1. & \begin{tabular}[l]{@{}l@{}}A lack of emotional expression, as if the person's \\ face is in a resting state. \end{tabular} \\
2. & \begin{tabular}[l]{@{}l@{}} The facial muscles are generally relaxed, creating a \\ smooth and even appearance. \end{tabular} \\
3. & \begin{tabular}[l]{@{}l@{}} The mouth is typically closed or slightly open, with \\ the lips not turned up or down. \end{tabular} \\
4. & \begin{tabular}[l]{@{}l@{}} The eyebrows are in a neutral position, not furrowed \\ or raised, and the eyes are generally looking straight ahead \\ or slightly down. \end{tabular} \\
5. & \begin{tabular}[l]{@{}l@{}} While the face may not show any specific emotions, \\ the expression can still convey a sense of attentiveness or alertness.  \end{tabular} \\
\hline\hline
\end{tabular}
\caption{Descriptions used for neutral category samples during training}
\label{tbl:neutral}
\end{table}

As AFEW~\cite{afew}, DFEW~\cite{dfew} and FERV39K~\cite{ferv39k} only have seven emotional classes, i.e. \textit{anger, disgust, fear, happiness, neutral, sadness \& surprise}, during inference we exclude descriptions for the additional classes (\textit{contempt, anxiety, helplessness \& disappointment}).

\section*{Qualitative Analysis}
\label{sec:qual}

Figure \ref{fig:quality} showcases a selection of correctly and incorrectly classified instances from the MAFW dataset. As the interpretation of emotion for humans is dependent on not only the facial expression but also the context, we see that some samples are harder than others to classify. It appears that examples with higher emotional intensity and more animated subjects appear to be easier to classify. Conversely, subtle expressions of emotion are prone to misclassification. For instance, in the anger examples depicted in Figure \ref{fig:quality}, the man displays multiple anger-associated expressions, such as a furrowed brow, whereas the woman exhibits a calmer demeanour. Similarly, in the happiness examples, the appropriately classified sample features a smiling man, while the incorrectly classified example shows a man with a subtle facial expression while speaking.

{\small
\bibliographystyle{ieee}

\begin{thebibliography}{10}\itemsep=-1pt

\bibitem{alayrac2022flamingo}
J.-B. Alayrac, J.~Donahue, P.~Luc, A.~Miech, I.~Barr, Y.~Hasson, K.~Lenc, A.~Mensch, K.~Millican, M.~Reynolds, R.~Ring, E.~Rutherford, S.~Cabi, T.~Han, Z.~Gong, S.~Samangooei, M.~Monteiro, J.~Menick, S.~Borgeaud, A.~Brock, A.~Nematzadeh, S.~Sharifzadeh, M.~Binkowski, R.~Barreira, O.~Vinyals, A.~Zisserman, and K.~Simonyan.
\newblock Flamingo: a visual language model for few-shot learning.
\newblock In {\em Advances in Neural Information Processing Systems}, 2022.

\bibitem{bagad_test_2023}
P.~Bagad, M.~Tapaswi, and C.~G.~M. Snoek.
\newblock Test of {Time}: {Instilling} {Video}-{Language} {Models} with a {Sense} of {Time}.
\newblock In {\em IEEE/CVF Conference on Computer Vision and Pattern Recognition (CVPR)}, Apr. 2023.

\bibitem{bain_clip-hitchhikers_2022}
M.~Bain, A.~Nagrani, G.~Varol, and A.~Zisserman.
\newblock A {CLIP}-{Hitchhiker}'s {Guide} to {Long} {Video} {Retrieval}, May 2022.

\bibitem{banerjee_learning_2022}
A.~Banerjee, U.~Bhattacharya, and A.~Bera.
\newblock Learning {Unseen} {Emotions} from {Gestures} via {Semantically}-{Conditioned} {Zero}-{Shot} {Perception} with {Adversarial} {Autoencoders}.
\newblock {\em Proceedings of the AAAI Conference on Artificial Intelligence}, 36(1):3--10, June 2022.
\newblock Number: 1.

\bibitem{bishay2019schinet}
M.~Bishay, P.~Palasek, S.~Priebe, and I.~Patras.
\newblock Schinet: Automatic estimation of symptoms of schizophrenia from facial behaviour analysis.
\newblock {\em IEEE Transactions on Affective Computing}, 12(4):949--961, 2019.

\bibitem{chowdary2021deep}
M.~K. Chowdary, T.~N. Nguyen, and D.~J. Hemanth.
\newblock Deep learning-based facial emotion recognition for human--computer interaction applications.
\newblock {\em Neural Computing and Applications}, pages 1--18, 2021.

\bibitem{cowen_self-report_2017}
A.~S. Cowen and D.~Keltner.
\newblock Self-report captures 27 distinct categories of emotion bridged by continuous gradients.
\newblock {\em Proceedings of the National Academy of Sciences}, 114(38):E7900--E7909, 2017.

\bibitem{afew}
A.~Dhall, R.~Goecke, S.~Lucey, T.~Gedeon, et~al.
\newblock Collecting large, richly annotated facial-expression databases from movies.
\newblock {\em IEEE multimedia}, 19(3):34, 2012.

\bibitem{ekman_facial_1978}
P.~Ekman and W.~V. Friesen.
\newblock {\em Facial action coding system: {Investigator}'s guide}.
\newblock Consulting Psychologists Press, 1978.

\bibitem{forbes_initial_2010}
C.~Forbes, J.~J. Blanchard, M.~Bennett, W.~P. Horan, A.~Kring, and R.~Gur.
\newblock Initial development and preliminary validation of a new negative symptom measure: {The} {Clinical} {Assessment} {Interview} for {Negative} {Symptoms} ({CAINS}).
\newblock {\em Schizophrenia Research}, 124(1-3):36--42, Dec. 2010.

\bibitem{foteinopoulou_learning_2022}
N.~M. Foteinopoulou and I.~Patras.
\newblock Learning from {Label} {Relationships} in {Human} {Affect}.
\newblock In {\em Proceedings of the 30th {ACM} {International} {Conference} on {Multimedia}}, pages 80--89, Lisboa Portugal, Oct. 2022. ACM.

\bibitem{foteinopoulou_estimating_2021}
N.~M. Foteinopoulou, C.~Tzelepis, and I.~Patras.
\newblock Estimating continuous affect with label uncertainty.
\newblock In {\em 2021 9th {International} {Conference} on {Affective} {Computing} and {Intelligent} {Interaction} ({ACII})}, pages 1--8, Nara, Japan, Sept. 2021. IEEE.

\bibitem{jia2021align}
C.~Jia, Y.~Yang, Y.~Xia, Y.-T. Chen, Z.~Parekh, H.~Pham, Q.~Le, Y.-H. Sung, Z.~Li, and T.~Duerig.
\newblock Scaling up visual and vision-language representation learning with noisy text supervision.
\newblock In {\em International Conference on Machine Learning}, pages 4904--4916. PMLR, 2021.

\bibitem{dfew}
X.~Jiang, Y.~Zong, W.~Zheng, C.~Tang, W.~Xia, C.~Lu, and J.~Liu.
\newblock Dfew: A large-scale database for recognizing dynamic facial expressions in the wild.
\newblock In {\em Proceedings of the 28th ACM international conference on multimedia}, pages 2881--2889, 2020.

\bibitem{kay_positive_1987}
S.~R. Kay, A.~Fiszbein, and L.~A. Opler.
\newblock The {Positive} and {Negative} {Syndrome} {Scale} ({PANSS}) for {Schizophrenia}.
\newblock {\em Schizophrenia Bulletin}, 13(2):261--276, Jan. 1987.

\bibitem{li_cliper_2023}
H.~Li, H.~Niu, Z.~Zhu, and F.~Zhao.
\newblock {CLIPER}: {A} {Unified} {Vision}-{Language} {Framework} for {In}-the-{Wild} {Facial} {Expression} {Recognition}, Feb. 2023.
\newblock arXiv:2303.00193 [cs].

\bibitem{li2022nr}
H.~Li, M.~Sui, Z.~Zhu, et~al.
\newblock Nr-dfernet: Noise-robust network for dynamic facial expression recognition.
\newblock {\em arXiv preprint arXiv:2206.04975}, 2022.

\bibitem{li_fer-former_2023}
Y.~Li, M.~Wang, M.~Gong, Y.~Lu, and L.~Liu.
\newblock {FER}-former: {Multi}-modal {Transformer} for {Facial} {Expression} {Recognition}, Mar. 2023.
\newblock arXiv:2303.12997 [cs].

\bibitem{lin_frozen_2022}
Z.~Lin, S.~Geng, R.~Zhang, P.~Gao, G.~de~Melo, X.~Wang, J.~Dai, Y.~Qiao, and H.~Li.
\newblock Frozen {CLIP} {Models} are {Efficient} {Video} {Learners}.
\newblock In S.~Avidan, G.~Brostow, M.~Cissé, G.~M. Farinella, and T.~Hassner, editors, {\em Computer {Vision} – {ECCV} 2022}, Lecture {Notes} in {Computer} {Science}, pages 388--404, Cham, 2022. Springer Nature Switzerland.

\bibitem{liu_mafw_2022}
Y.~Liu, W.~Dai, C.~Feng, W.~Wang, G.~Yin, J.~Zeng, and S.~Shan.
\newblock {MAFW}: {A} {Large}-scale, {Multi}-modal, {Compound} {Affective} {Database} for {Dynamic} {Facial} {Expression} {Recognition} in the {Wild}.
\newblock In {\em Proceedings of the 30th {ACM} {International} {Conference} on {Multimedia}}, {MM} '22, pages 24--32, New York, NY, USA, Oct. 2022. Association for Computing Machinery.

\bibitem{luo_clip4clip_2022}
H.~Luo, L.~Ji, M.~Zhong, Y.~Chen, W.~Lei, N.~Duan, and T.~Li.
\newblock {CLIP4Clip}: {An} empirical study of {CLIP} for end to end video clip retrieval and captioning.
\newblock {\em Neurocomputing}, 508:293--304, Oct. 2022.

\bibitem{ma_x-clip_2022}
Y.~Ma, G.~Xu, X.~Sun, M.~Yan, J.~Zhang, and R.~Ji.
\newblock X-{CLIP}: {End}-to-{End} {Multi}-grained {Contrastive} {Learning} for {Video}-{Text} {Retrieval}.
\newblock In {\em Proceedings of the 30th {ACM} {International} {Conference} on {Multimedia}}, {MM} '22, pages 638--647. Association for Computing Machinery, Oct. 2022.

\bibitem{menon2022visual}
S.~Menon and C.~Vondrick.
\newblock Visual classification via description from large language models.
\newblock {\em ICLR}, 2023.

\bibitem{ouali_black_2023}
Y.~Ouali, A.~Bulat, B.~Martinez, and G.~Tzimiropoulos.
\newblock Black {Box} {Few}-{Shot} {Adaptation} for {Vision}-{Language} models, Apr. 2023.

\bibitem{parisot_learning_2023}
S.~Parisot, Y.~Yang, and S.~McDonagh.
\newblock Learning to {Name} {Classes} for {Vision} and {Language} {Models}.
\newblock In {\em IEEE/CVF Conference on Computer Vision and Pattern Recognition (CVPR)}, Apr. 2023.

\bibitem{ness}
S.~Priebe, M.~Savill, T.~Wykes, R.~Bentall, U.~Reininghaus, C.~Lauber, S.~Bremner, S.~Eldridge, and F.~Röhricht.
\newblock Effectiveness of group body psychotherapy for negative symptoms of schizophrenia: multicentre randomised controlled trial.
\newblock {\em The British Journal of Psychiatry}, 209(1):54--61, 2016.

\bibitem{qi_zero-shot_2021}
F.~Qi, X.~Yang, and C.~Xu.
\newblock Zero-shot {Video} {Emotion} {Recognition} via {Multimodal} {Protagonist}-aware {Transformer} {Network}.
\newblock In {\em Proceedings of the 29th {ACM} {International} {Conference} on {Multimedia}}, {MM} '21, pages 1074--1083, New York, NY, USA, Oct. 2021. Association for Computing Machinery.

\bibitem{radford_CLIP_2021}
A.~Radford, J.~W. Kim, C.~Hallacy, A.~Ramesh, G.~Goh, S.~Agarwal, G.~Sastry, A.~Askell, P.~Mishkin, J.~Clark, G.~Krueger, and I.~Sutskever.
\newblock Learning transferable visual models from natural language supervision.
\newblock In M.~Meila and T.~Zhang, editors, {\em Proceedings of the 38th International Conference on Machine Learning}, volume 139 of {\em Proceedings of Machine Learning Research}, pages 8748--8763. PMLR, 18--24 Jul 2021.

\bibitem{russell_circumplex_1980}
J.~A. Russell.
\newblock A circumplex model of affect.
\newblock {\em Journal of Personality and Social Psychology}, 39(6):1161--1178, 1980.

\bibitem{laion}
C.~Schuhmann, R.~Vencu, R.~Beaumont, R.~Kaczmarczyk, C.~Mullis, A.~Katta, T.~Coombes, J.~Jitsev, and A.~Komatsuzaki.
\newblock {LAION}-{400M}: {Open} {Dataset} of {CLIP}-{Filtered} 400 {Million} {Image}-{Text} {Pairs}, Nov. 2021.
\newblock arXiv:2111.02114 [cs].

\bibitem{tang_clip4caption_2021}
M.~Tang, Z.~Wang, Z.~LIU, F.~Rao, D.~Li, and X.~Li.
\newblock {CLIP4Caption}: {CLIP} for {Video} {Caption}.
\newblock In {\em Proceedings of the 29th {ACM} {International} {Conference} on {Multimedia}}, {MM} '21, pages 4858--4862, New York, NY, USA, Oct. 2021. Association for Computing Machinery.

\bibitem{tron_automated_2015}
T.~Tron, A.~Peled, A.~Grinsphoon, and D.~Weinshall.
\newblock Automated facial expressions analysis in schizophrenia: {A} continuous dynamic approach.
\newblock In {\em International {Symposium} on {Pervasive} {Computing} {Paradigms} for {Mental} {Health}}, pages 72--81. Springer, 2015.

\bibitem{tron_facial_2016}
T.~Tron, A.~Peled, A.~Grinsphoon, and D.~Weinshall.
\newblock Facial expressions and flat affect in schizophrenia, automatic analysis from depth camera data.
\newblock In {\em 2016 {IEEE}-{EMBS} {International} {Conference} on {Biomedical} and {Health} {Informatics} ({BHI})}, pages 220--223. IEEE, 2016.

\bibitem{wang_actionclip_2021}
M.~Wang, J.~Xing, and Y.~Liu.
\newblock {ActionCLIP}: {A} {New} {Paradigm} for {Video} {Action} {Recognition}, Sept. 2021.
\newblock arXiv:2109.08472 [cs].

\bibitem{ferv39k}
Y.~Wang, Y.~Sun, Y.~Huang, Z.~Liu, S.~Gao, W.~Zhang, W.~Ge, and W.~Zhang.
\newblock Ferv39k: a large-scale multi-scene dataset for facial expression recognition in videos.
\newblock In {\em Proceedings of the IEEE/CVF Conference on Computer Vision and Pattern Recognition}, pages 20922--20931, 2022.

\bibitem{xian2018zero}
Y.~Xian, C.~H. Lampert, B.~Schiele, and Z.~Akata.
\newblock Zero-shot learning—a comprehensive evaluation of the good, the bad and the ugly.
\newblock {\em IEEE transactions on pattern analysis and machine intelligence}, 41(9):2251--2265, 2018.

\bibitem{xu_exploring_2022}
X.~Xu, J.~Deng, N.~Cummins, Z.~Zhang, L.~Zhao, and B.~W. Schuller.
\newblock Exploring {Zero}-{Shot} {Emotion} {Recognition} in {Speech} {Using} {Semantic}-{Embedding} {Prototypes}.
\newblock 24, 2022.
\newblock Conference Name: IEEE Transactions on Multimedia.

\bibitem{xu2023zero}
X.~Xu, J.~Deng, Z.~Zhang, Z.~Yang, and B.~W. Schuller.
\newblock Zero-shot speech emotion recognition using generative learning with reconstructed prototypes.
\newblock In {\em ICASSP 2023-2023 IEEE International Conference on Acoustics, Speech and Signal Processing (ICASSP)}, pages 1--5. IEEE, 2023.

\bibitem{xue_clip-vip_2022}
H.~Xue, Y.~Sun, B.~Liu, J.~Fu, R.~Song, H.~Li, and J.~Luo.
\newblock {CLIP}-{ViP}: {Adapting} {Pre}-trained {Image}-{Text} {Model} to {Video}-{Language} {Representation} {Alignment}, Sept. 2022.
\newblock arXiv:2209.06430 [cs].

\bibitem{YADEGARIDEHKORDI2019103649}
E.~Yadegaridehkordi, N.~F. B.~M. Noor, M.~N.~B. Ayub, H.~B. Affal, and N.~B. Hussin.
\newblock Affective computing in education: A systematic review and future research.
\newblock {\em Computers \& Education}, 142:103649, 2019.

\bibitem{yang_vid2seq_2023}
A.~Yang, A.~Nagrani, P.~H. Seo, A.~Miech, J.~Pont-Tuset, I.~Laptev, J.~Sivic, and C.~Schmid.
\newblock {Vid2Seq}: {Large}-{Scale} {Pretraining} of a {Visual} {Language} {Model} for {Dense} {Video} {Captioning}, Feb. 2023.
\newblock arXiv:2302.14115 [cs].

\bibitem{yuksekgonul_when_2023}
M.~Yuksekgonul, F.~Bianchi, P.~Kalluri, D.~Jurafsky, and J.~Zou.
\newblock When and why {Vision}-{Language} {Models} behave like {Bags}-of-{Words}, and what to do about it?
\newblock In {\em International {Conference} on {Learning} {Representations}}, 2023.

\bibitem{zara_autolabel_2023}
G.~Zara, S.~Roy, P.~Rota, and E.~Ricci.
\newblock {AutoLabel}: {CLIP}-based framework for {Open}-set {Video} {Domain} {Adaptation}, Apr. 2023.
\newblock arXiv:2304.01110 [cs].

\bibitem{zhan_zero-shot_2019}
C.~Zhan, D.~She, S.~Zhao, M.-M. Cheng, and J.~Yang.
\newblock Zero-shot emotion recognition via affective structural embedding.
\newblock In {\em Proceedings of the IEEE/CVF International Conference on Computer Vision}, pages 1151--1160, 2019.

\bibitem{zheng_FARL_2022}
Y.~Zheng, H.~Yang, T.~Zhang, J.~Bao, D.~Chen, Y.~Huang, L.~Yuan, D.~Chen, M.~Zeng, and F.~Wen.
\newblock General {Facial} {Representation} {Learning} in a {Visual}-{Linguistic} {Manner}.
\newblock In {\em 2022 {IEEE}/{CVF} {Conference} on {Computer} {Vision} and {Pattern} {Recognition} ({CVPR})}, pages 18676--18688, June 2022.

\bibitem{zheng2022general}
Y.~Zheng, H.~Yang, T.~Zhang, J.~Bao, D.~Chen, Y.~Huang, L.~Yuan, D.~Chen, M.~Zeng, and F.~Wen.
\newblock General facial representation learning in a visual-linguistic manner.
\newblock In {\em Proceedings of the IEEE/CVF Conference on Computer Vision and Pattern Recognition}, pages 18697--18709, 2022.

\bibitem{zhou_conditional_2022}
K.~Zhou, J.~Yang, C.~C. Loy, and Z.~Liu.
\newblock Conditional prompt learning for vision-language models.
\newblock In {\em IEEE/CVF Conference on Computer Vision and Pattern Recognition (CVPR)}, 2022.

\bibitem{zhou_learning_2022}
K.~Zhou, J.~Yang, C.~C. Loy, and Z.~Liu.
\newblock Learning to prompt for vision-language models.
\newblock {\em International Journal of Computer Vision (IJCV)}, 2022.

\end{thebibliography}

}

\end{document}